\documentclass[conference]{IEEEtran}
\IEEEoverridecommandlockouts
\usepackage{cite}
\usepackage{amsmath,amssymb,amsfonts}
\usepackage{algorithmic}
\usepackage{graphicx}
\usepackage{textcomp}
\usepackage{xcolor}
\usepackage[colorlinks=true,linkcolor=blue]{hyperref}%
\usepackage{graphicx}
\usepackage{amssymb}
\def\BibTeX{{\rm B\kern-.05em{\sc i\kern-.025em b}\kern-.08em
    T\kern-.1667em\lower.7ex\hbox{E}\kern-.125emX}}
\begin{document}

\title{Multi-Modal Zero-Shot Sign Language Recognition\\}


\author{Razieh Rastgoo$^{1,2}$, Kourosh Kiani$^1$,Sergio Escalera$^3$, Mohammad Sabokrou$^2$  \\
$^1$Semnan University~~ $^2$Institute for Research in Fundamental Sciences (IPM)\\
~~ $^3$Universitat de Barcelona and Computer Vision Center\\
}

\maketitle

\begin{abstract}
Zero-Shot Learning (ZSL) has rapidly advanced in recent years. Towards overcoming the annotation bottleneck in the Sign Language Recognition (SLR), we explore the idea of Zero-Shot Sign Language Recognition (ZS-SLR) with no annotated visual examples, by leveraging their textual descriptions. In this way, we propose a multi-modal Zero-Shot Sign Language Recognition (ZS-SLR) model harnessing from the complementary capabilities of deep features fused with the skeleton-based ones. A Transformer-based model along with a C3D model is used for hand detection and deep features extraction, respectively. To make a trade-off between the dimensionality of the skeleton-based and deep features, we use an Auto-Encoder (AE) on top of the Long Short Term Memory (LSTM) network. Finally, a semantic space is used to map the visual features to the lingual embedding of the class labels, achieved via the Bidirectional Encoder Representations from Transformers (BERT) model. Results on four large-scale datasets, RKS-PERSIANSIGN, First-Person, ASLVID, and isoGD, show the superiority of the proposed model compared to state-of-the-art alternatives in ZS-SLR.
\end{abstract}


\section{Introduction}
Advances in Sign Language Recognition (SLR) have been predominantly driven by the advent of accurate technologies. Furthermore, deep neural networks have achieved the best performance in most of vision tasks \cite{b6,b7,color,face}, such as SLR \cite{b00,rbm,b1,b2,b3,b4,b5}, video classification \cite{bb8}, action recognition \cite{bb9}, and gesture recognition \cite{bb10,infosci3}. The necessity to a huge number of labelled training samples makes such model inherently biased towards the prediction phase, and also less useful where there might not be enough labeled data for all classes. In addition, the conventional deep learning methods for SLR are not able to recognize the samples from a new class/concept. This is where Zero-Shot Learning (ZSL) comes to use. To this end, ZSL evaluates the effectiveness of the embedding space, which is made using the input data and the auxiliary information.

Although the previous SLR methods have achieved the state-of-the-art performance \cite{b1,b2,b3,b4,b5}, they suffer from the annotation bottleneck and do not work efficiently if face with a sample from an unseen class in training duration. To address the mentioned weaknesses, we explore the idea of ZS-SLR with no annotated visual examples. 
We propose a ZSL model for SLR from multi-modal data and hybrid features. One of the input modalities used in the proposed model is the 3D skeleton data that has also received considerable attention in recent years \cite{b1,b2,b3,b4,b5}. Generally, skeleton representation is beneficial because it is compact and can robustly separate the action subject (human) from the background. Furthermore, the presentation of large-scale skeleton annotated datasets makes a room for researchers to develop different approaches for skeleton-based SLR. Considering these advantages, the proposed model in this work employs some skeleton-based features.

While ZSL plays an effective role to get closer to real-life applications, there are some major challenges in ZSL, including domain shift, bias, cross-domain knowledge transfer, semantic loss, and hubness. To tackle these challenges, there are two main methodologies used in ZSL-based tasks: embedding-based and generative-based models. While embedding-based models aim to map the visual features and semantic attributes into a common embedding space, generative-based models target to generate visual features for unseen classes using the semantic attributes. In this paper, we propose an embedding-based model using Transformer \cite{Transformer-Object}, 3D Convolutional Neural Network (3DCNN) \cite{C3D}, Auto-Encoder (AE) \cite{AE}, Long Short Term Memory (LSTM) \cite{LSTM}, and Bidirectional Encoder Representations from Transformers (BERT) model \cite{BERT} from multi-modal inputs and hybrid features for ZS-SLR.

Our main contributions can be listed as follows:

\textbf{ZS-SLR: } Towards overcoming the annotation bottleneck in the SLR, we formulate the problem of ZS-SLR with no annotated visual examples, by leveraging their textual descriptions.

\textbf{Hand detection: } Since hand detection is an important step for hand sign language recognition, we configure a Transformer-based model, as a fast and accurate hand detection model, to tackle the challenges of the current hand detection models. To the best of our knowledge, this is the first time that a Transformer-based model is configured for the hand detection.

\textbf{Hybrid features representation:} We propose a handcrafted plus deep end-to-end hybrid model for ZS-SLR. We use some skeleton-based features, including distances, angles, and singular values of the human hand joints, fused with deep features. To the best of our knowledge, this is the first time that skeleton-based features and deep features are working together within ZS-SLR.

\textbf{Multi-modal inputs:} We include three modalities in the model: skeleton, RGB video, and text. Using the complementary capabilities of three modalities is beneficial for our model. The proposed model is the first multi-modal model for the ZS-SLR task.

\textbf{Performance:} We perform a detailed analysis of the proposed model using two evaluation protocols. Results on four large-scale datasets, RKS-PERSIANSIGN, First-Person, ASLVID, and isoGD, show the superiority of the proposed model compared to the state-of-the-art alternatives in the ZS-SLR.

The remainder of this paper is organized as follows. Section 2 briefly reviews recent works in the ZS-SLR, Zero-Shot Gesture Recognition (ZS-GR), and Zero-Shot Action Recognition (ZS-AR). The proposed model is presented in details in section 3. Results are discussed in section 4. Section 5 and 6 analyze the results and conclude the work with the comments on possible lines for future research, respectively.\\
 
\section{Related work}
The ZSL idea was initially proposed by Palatucci et al. \cite{b8} and Larochelle et al. \cite{b9}. Since there is only one work in ZS-SLR \cite{Bilge}, we briefly review recent works in the related areas, especially ZS-GR and ZS-AR (Summarized in Table \ref{Table 1}). The ZSL scenario typically needs to map the visual features to the semantic embedding obtained from the unseen data\cite{Kodi,Wang,Xu}. While the semantic representation can be either class-related attributes or the embedding of the class labels \cite{Liu,Mishra}, the visual representation is either handcrafted features \cite{Qin,Xun,Yi} based on the Improved Dense Trajectories (IDT) method \cite{Heng}, or deep \cite{Gan,Mishra,Wang} features extracted with pre-trained models, such as C3D network \cite{C3D}. While the IDT features can be represented using a single vector, deep features refer to a predefined length of a video segment. Different techniques can be applied to the semantic and visual domains to obtain a powerful discriminative capability in the ZSL models coping with unseen data. Generally, we can categorize the reviewed work into Inductive and Transductive ZSL:

\textbf{Inductive ZSL:} In this group, only the labeled training data from seen classes are accessible and the test data is completely unknown at training time. Bishay et al. proposed a model, entitled Temporal Attentive Relation Network (TARN), for ZS-AR. A combination of a C3D network and Bidirectional Gated Recurrent Unit (Bi-GRU) is used to obtain a single vector from the embedding module. Finally, the encoded video is mapped into Word2Vec embedding. Results on the UCF101 and HMDB-51datasets show that the model presents a comparable performance with the state-of-the-art alternatives in ZS-AR \cite{Bishay}. Hahn et al. proposed a model, entitled Action2Vec, by combining linguistic embedding of the class labels with spatio-temporal features obtained from the video inputs. The architecture of this model includes a C3D model for visual features extraction and a two-layer hierarchical LSTM network for temporal features. Results on the UCF101, HMDB-51, and Kinetics datasets confirm the superiority of the Action2Vec model, achieving state-of-the-art with 1.3\%\, 4.38\%\, and 7.75\%\ relative margins, respectively \cite{Hahn}. Gupta et al. proposed a generative-based model for ZS-AR from the skeleton data. The Variational Auto Encoder (VAE) is used as the base architecture to learn the generative space of the latent representations. The latent visual representations of skeleton-based human actions are fused with the syntactic information obtained from the textual descriptions. Results on the NTU-60 and NTU-120 datasets show the outperforming of the model performance compared to state-of-the-art models in ZS-AR with the 4.34\%\ and 3.16\%\ relative improvements, respectively \cite{b6}. Bilge et al. proposed a model for ZS-SLR using the combination of I3D and BERT. The body and hand regions are used to obtain the visual features through 3D-CNNs. The longer temporal relationships are obtained via Bidirectional Long-Short-Term Memory (BLSTM) network. Relying on the textual descriptions, they extended the current ASL dataset, namely ASL-Text, which includes 250 signs and the corresponding sign descriptions. Results on this dataset show that this model can provide a basis for further exploration of the ZSL in SLR \cite{Bilge}. Madapana and Wachs proposed a bi-linear Auto-Encoder (AE) model for ZS-GR that jointly optimizes reconstruction and classification errors. Furthermore, they analyzed three feature extraction techniques, raw features, handcrafted features, and deep features, and conducted experiments to compare unseen class accuracies obtained using these methods. Results on a dataset, including a subset of two public datasets, show that the model achieves comparable results with state-of-the-art models in the ZS-GR \cite{Mada}. Wu et al. proposed an attribute-based model for zero-shot dynamic hand gestures recognition using a BLSTM network. The skeletal joint data obtained by Leap Motion Controller (LMC) is used in the model. Relying on the extracted features from the BLSTM network and semantic attributes, a Semantic Auto-encoder (SAE) is used to learn a mapping from feature space to semantic space. Results on own dataset show the effectiveness of the proposed model \cite{Wu18}.\\
\textbf{Transductive ZSL:} In this group, the unlabeled test data are available during training. Alexiou et al. proposed a ZSL model using Fisher Vector (FV) method, as a low-complex and unsupervised method, to obtain the visual representation of the input video. A Support Vector Machine (SVM) regressor is used to make a mapping from the visual domain to the word-vector domain. Furthermore, they employ a search algorithm to find a set of words with similar semantic meanings to the class labels. Results on two large-scale datasets, UCF101 and HMDB-51, show a recognition accuracy improvement compared to state-of-the-art models in ZS-AR \cite{Alex}. 
Mishra et al. provided a probabilistic generative model for ZS-AR by representing each class via a Gaussian distribution model. This model can synthesize new examples for any unseen class by sampling from the class distribution. The C3D network is used for visual features extraction. Results on three datasets, UCF101, HMDB-51, and Olympic, show that the proposed approach achieves a comparable performance with the state-of-the-art methods in the filed \cite{Mishra}. Wang and Chen proposed a ZS-AR model using textual descriptions as a word vector representation. Furthermore, they employed the action-related still images for semantic representation of the video-based human actions in ZSL. The visual features are extracted using a CNN model. Results on UCF101 and HMDB-51datasets show the effectiveness of the proposed semantic representations in the model \cite{Wu18}.\\

\begin{table*}[h!]
\thispagestyle{empty}
\caption{\label{Table 1} A summary of the reviewed works.}
\begin{center}
{\small
 \noindent\begin{tabular}{p{2cm}p{1.5cm}p{4cm}p{4cm}p{1.5cm}p{1cm}}
 \hline
\textbf{Method} & \textbf{Ref.} & \textbf{Model} & \textbf{Dataset} & \textbf{Task} & \textbf{Year} \\
\hline\hline
Inductive & \cite{Bishay} & C3D, Bi-GRU, Word2Vec & UCF101, HMDB-51 & AR & 2019\\
&	\cite{Hahn} & C3D, LSTM & UCF101, HMDB-51 & AR & 2019\\
&	\cite{b6} & VAE & NTU-60, NTU-120 & AR & 2021\\
&	\cite{Bilge} & I3D, BLSTM, BERT & ASL-Text & SLR & 2019\\
&	\cite{Mada} & BLSTM & Subset of CGD 2013 dataset and MSRC-12 & GR & 2020\\
\hline
Transductive & \cite{Alex}	& FV, SVM & UCF101, HMDB-51 & AR & 2016\\
&	\cite{Mishra}	& C3D, Word2Vec &	UCF101, HMDB-51, Olympic & AR &2018\\
&	\cite{Wu18}& CNN & UCF101, HMDB-51 & AR & 2017\\
\noalign{\smallskip}\hline
\end{tabular}
 }
\end{center}
\end{table*}

\section{Proposed model}
\begin{figure*}[htbp]
\centerline{\includegraphics[width=0.8\linewidth]{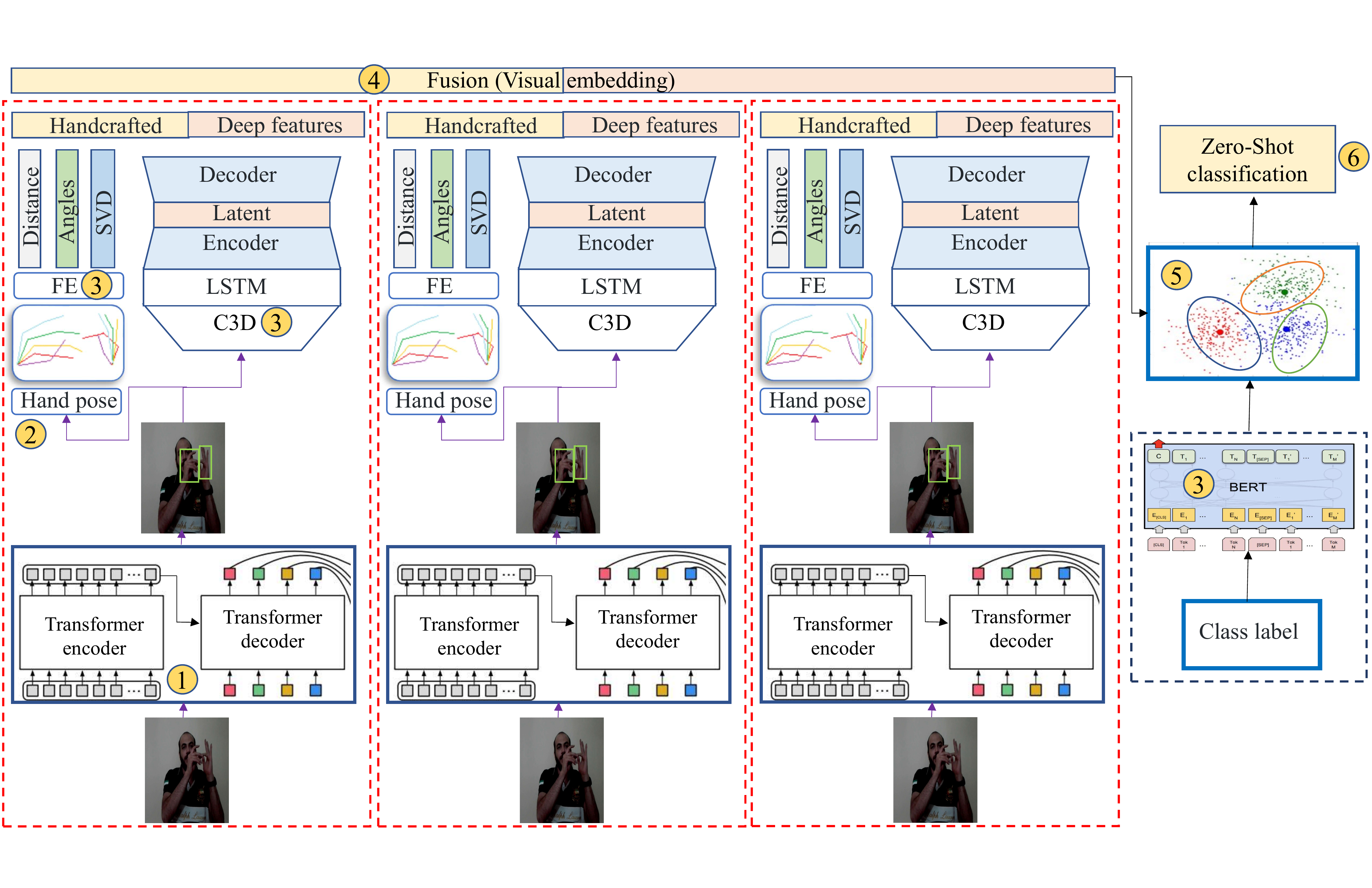}}
\caption{Proposed model for ZS-SLR, including six main blocks: (1) Transformer-based hand detection, (2) Hand pose estimation, (3) Lingual, skeleton-based, and deep feature extraction, (4) Features fusion, (5) Semantic space, and (6) Classification.}
\label{fig 1}
\end{figure*}

In this section, we present the details of the proposed model. Figure \ref{fig 1} shows an overview of the proposed architecture.

\subsection{Problem definition}
Let $V_{S} = \{(x_{S}^{1},y_{S}^{1}),(x_{S}^{2},y_{S}^{2}),...,(x_{S}^{N},y_{S}^{N})\}$ denote the set of N pairs of video x and the corresponding class label y of the seen data during the training with the subscript S standing for seen data. On similar lines, $V_{S} = \{(x_{U}^{1},y_{U}^{1}),(x_{U}^{2},y_{U}^{2}),...,(x_{S}^{M},y_{S}^{M})\}$  denote the set of M pairs of video x and the corresponding class label y of the unseen data during the testing with the subscript U standing for unseen data. Suppose $y_{S}\in Y_{S}, y_{U}\in Y_{U}$. Also, consider ${y}_{t}$  represents the test time class prediction. For ZS-SLR, we have ${y}_{t}\in Y_{U},Y_{S} {\bigcap} Y_{U} = \O$. ZSL classifiers require generalization capability to unseen test classes. One way to have this capability is using nearest-neighbor search in a semantic space. To make an inference, given a video $x_{U}^{i}$, we infer the corresponding semantic embedding $z = g(x_{U}^{i})$ and classify  as the nearest neighbor of z in the embedding of the test classes. Finally, a trained classification model M(·) outputs
\begin{equation}
    M(x_{U}^{i}) = \underset{y_{U} \in Y_{U}}{\mathrm{argmin} \cos{(g(x_{U}^{i}),BERT(y_{U})})},
\end{equation}
where $\cos$ is the cosine distance and the semantic embedding is computed using BERT \cite{BERT}, $BERT: y_{U} \to R^{1024}$. The function $g = f_{s} \circ f_{v}$ is a combination of the visual and lingual encoding. 

\subsection{Model details}
Different blocks of the proposed model are described in the following.

\subsubsection{Inputs}
Three input modalities are used in the model: skeleton, video, and text. Two input modalities, skeleton and video, are used for visual feature extraction. The text modality is employed for lingual embedding. The visual features and lingual embedding are used as the input and output of the semantic space, respectively. 

\subsubsection{Transformer-based hand detection}
Hand detection is an important step of the proposed model. While different models are used for hand detection, such as Faster-RCNN \cite{F-RCNN}, Single Shot Detector (SSD) \cite{SSD}, or You Only Look Once (YOLO) v3 \cite{YOLO}, the performance of these models is highly dependent on the hand-designed components like a Non-Maximum Suppression (NMS) procedure or anchor generation. These components need to explicitly encode the prior knowledge about the task. To tackle these challenges, we configure the DEtection TRansformer (DETE) model, as a Transformer-based model for object detection task developed by Facebook AI \cite{Transformer-Object}, for hand detection. DETR includes an end-to-end architecture by eliminating any customized layers to predict the bounding boxes. This model is configured for hand detection in the proposed model to benefit from the Transformer capabilities. 

\subsubsection{Hand pose estimation}
After hand detection, the 3D hand keypoints are obtained using the OpenPose model \cite{OpenPose}. This model estimates 21 3D hand keypoints for each hand. We will process the pixel regions of the detected hands and the estimated 3D hand keypoints, in the next block.  

\subsubsection{Features extraction} 
\begin{figure}[htbp]
\centerline{\includegraphics[width=0.8\linewidth]{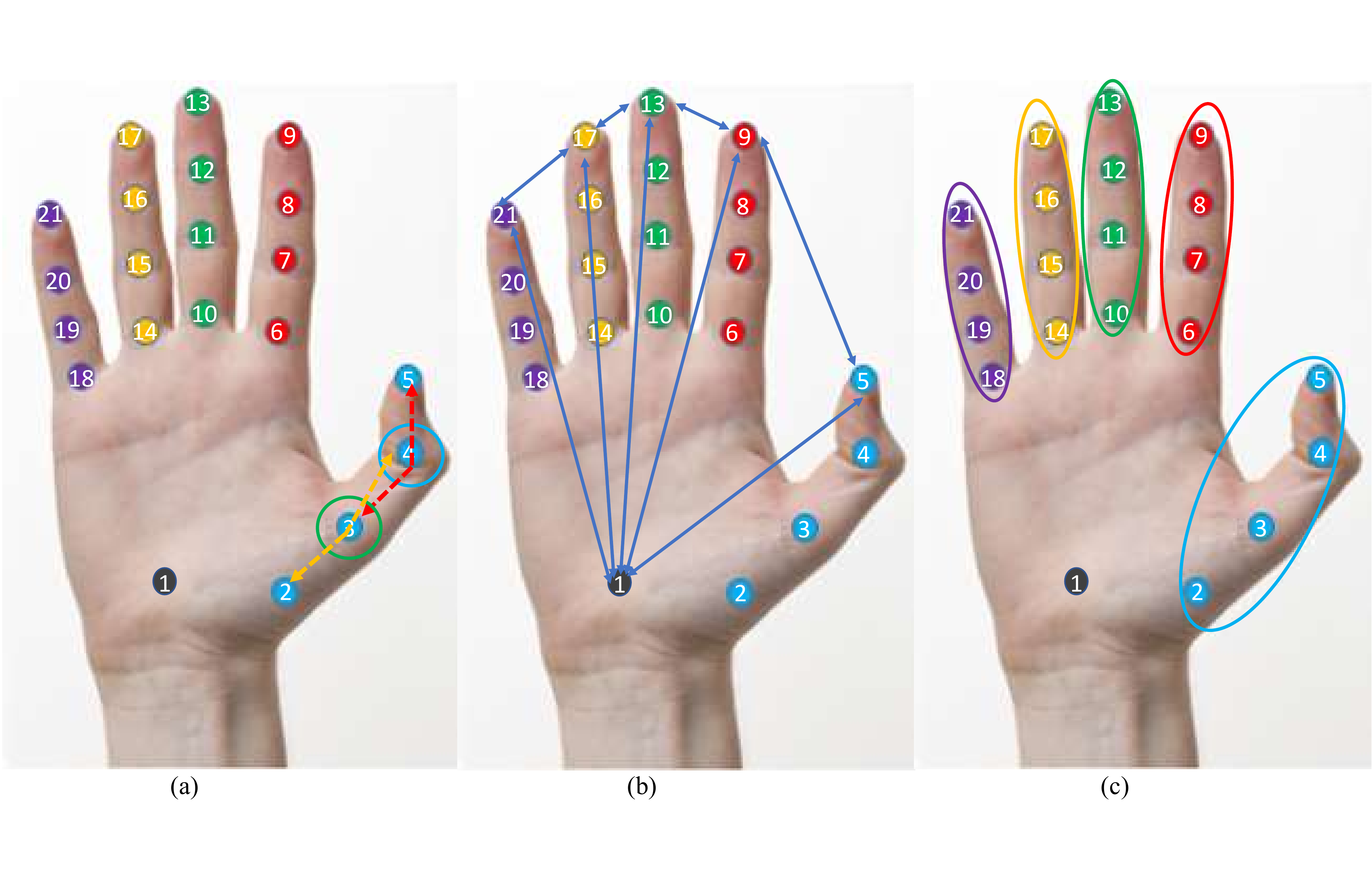}}
\caption{The skeleton-based features visualization: (a) Angle, (b): Distance, and (c) SVD.}
\label{fig 2}
\end{figure}

The features used in the proposed model fall into visual and lingual categories:\\
\textbf{Visual features:} In this block, the visual features are extracted from two main categories:\\
\begin{itemize}
    \item \textbf{Skeleton-based features:} Three feature types are extracted from skeleton data: distances, angles, and singular values from Singular Value Decomposition (SVD).\\ 
    \textbf{Distance features:} After exploring the relations between different joints of the human hands, some joint distances have been included. The norm value of the distance between two joints is included in the features. The middle part of the Figure \ref{fig 2} shows some samples of these features.\\
    \textbf{Angle features:} Relying on the relations between different joints of the hand fingers, we selected two consecutive triple sets of hand keypoints in each hand to obtain a more accurate description from them. In more detail, let $Set_{Keypoints} = \{\underbrace{(x_{1},y_{1},z_{1})}_{First},\underbrace{(x_{2},y_{2},z_{2})}_{Second},\underbrace{(x_{3},y_{3},z_{3})}_{Third},\underbrace{(x_{4},y_{4},z_{4})}_{Fourth}\}$ denote the set of four 3D keypoints of each hand finger. We compute two angles, $\theta_{1}$ and $\theta_{2}$, as the angles formed between two consecutive triple sets of these keypoints.
    
    
     The left part of the Figure \ref{fig 2} shows a sample of this feature.\\
    \textbf{SVD:} Based on the relations between different joints of the human hands, we selected the sets of four human hand keypoints to obtain the SVD features. 
    In more detail, considering $Set_{Keypoints}$, we calculate the SVD features and obtain $U,V,S$. We include the singular values, $V$, in the model features. The right part of the Figure \ref{fig 2} shows a sample of this feature.\\
    
    
    \item \textbf{Pixel-level features:} The C3D model is employed to extract the pixel-level features every 16 frames in the video. Then, the features of all snippets are passed to a LSTM layer on top of the C3D network to obtain the temporal features. Furthermore, to obtain a compact representation of the features, an Auto-Encoder (AE), including two Fully Connected (FC) layers, is stacked on the LSTM as a dimension reduction methodology. We apply an Auto-Encoder (AE) to the LSTM output obtained from video input to balance the dimentionality between the deep and skeleton-based features.\\
\end{itemize}
\textbf{Lingual features:} We use the sentence BERT model \cite{BERT} to obtain a 1024-dimensional word embedding.

\subsubsection{Features fusion} Three visual feature types are fused to input to the semantic space.\\
\subsubsection{Semantic space} The main goal of semantic space is to map the visual features to the lingual embedding using a projection function learned using deep networks. During training, the projection function is learned using seen data. Since deep networks can be used as a function approximator, the learned model is employed at the inference phase to predict the lingual embedding of unseen data. The predicted embedding achieved from the projection network is used to obtain the similarity degree to the unseen class embeddings. A simple similarity measure is using the embedding of the class labels. The similarity between the two classes is defined as follows:
\begin{equation}
    Similarity(c_{i},z_{1})= \cos{(BERT(c_{i}),z_{1})},
\end{equation}
where $c_{1}$ is the class label corresponding to the ith class from the unseen data and $z_{1}$ is the predicted class embedding using the trained projection network. This prediction is compared to all class embeddings obtained from the unseen data to select the closest one.\\
\subsubsection{Classification} In this phase, the final class label is recognized using a Softmax layer applied on the similarity degrees achieved from the previous phase.

\section{Results}
Here, the details of the model implementation and obtained results are presented.

\subsection{Implementation details}
The proposed model are evaluated on Intel(R) Xeon(R) CPU E5-2699 (2 processors) with 90GB RAM on Microsoft Windows 10 operating system and Python software on 10 NVIDIA GPUs. The PyTorch library is used for implementation. Four datasets are used for evaluation. All of the results are reported on the unseen test data. The proposed model is optimized using Adam with a learning rate for 1e-3 and a total training epoch of 300. Based on the C3D architecture, we use some video snippets, including 16 frames. For each video frame, we include three skeleton-based features, including the distances, angles, and singular values from the SVD. Since each sign can involve one or two hands, we define these skeleton-based features not only for one hand but also for the relations between two hands. In the signs containing only one hand, we simply repeat the features to be compatible with the case of having two hands. Details of these features are as follows (See Table \ref{Table 2}):\\
\textbf{Distance features:} We include 20 distance features for each hand in a sign. Furthermore, 21 distance features corresponding to the peer-to-peer distance between the hand joints are added to the individual distances for each hand. We totally have 61 distance features for each video sample.\\
\textbf{Angle features:} For each hand finger, we include two angle features. We have in total 10 angle features per hand and 20 angle features for two hands.\\
\textbf{SVD features:} While 21 3D hand keypoints are obtained for each hand, we use twenty of them and ignore the hand palm keypoint. Considering four keypoints per finger, 20 3D hand keypoints are placed in a matrix with a 4x15 shape. Furthermore, three singular values are obtained from the 21 3D hand keypoints. In this way, we have a total of seven singular values per hand. Considering the relation between two hands, we also apply the SVD method to the concatenated joints of two hands in four different shapes: 21x6, 42x3, 8x15, and 4x30. So, we have 21 singular values. We totally have 35 singular values for each video sample.\\
Considering all of the skeleton-based features, we have 116 features. The extracted features from the video snippets are input to the LSTM, including 1024 hidden neurons. Using an AE, we obtain a latent feature vector of 510 shape. To balance between the skeleton-based and deep features, we repeated the skeleton-based features before the fusion with deep features. Since the singular values play an important role in the skeleton-based features, we assign a higher weight to them. To this end, we repeat the distances, angles, and SVD features four, three, and six times, respectively. Since the shape of the pixel-level features is 510, the final fusion of the skeleton-based and deep features will get a feature vector with the 1024 shape. The lingual embedding outputs a 1024 embedding vector. In the semantic space, we have a deep model, including two Fully Connected (FC) layers to map the visual features into the lingual embedding. We use two evaluation protocols to analyze the results. These protocols are compatible with the current works in the field to make a fair comparison with state-of-the-art models.\\
\textbf{First evaluation protocol:} In this protocol, we randomly select 80\%\ of the classes for training and the other classes for testing. This protocol aims to prepare the model with more seen data.\\
\textbf{Second evaluation protocol:} In this protocol, we randomly select 50\%\ of the classes for training and the other classes for testing. This challenging protocol aims to assign equal weight to seen and unseen data.\\

\begin{table*}[h!]
\thispagestyle{empty}
\caption{\label{Table 2} Details of the skeleton-based features. Each brace shows a collection of the joints considered for a skeleton-based feature.}
\begin{center}
{\small
 \noindent\begin{tabular}{p{1.5cm}p{12cm}p{1.5cm}}
 \hline
\textbf{Feature} & \textbf{One-hand} & \textbf{Two-hands}\\
\hline\hline
Distance & (5,9), (9,13), (13,17), (17,21), (5,1), (9,1), (13,1), (17,1), (21,1), (5,2), (9,6), (13,10), (17,14), (21,18), (9,4), (9,12), (13,8), (13,16), (17,12), (17,20) & All joints\\
\hline
Angle & (2,3,4), (3,4,5), (6,7,8), (7,8,9), (10,11,12), (11,12,13), (14,15,16), (15,16,17), (18,19,20), (19,20,21) & -\\
\hline
SVD	& (2,3,4,5), (6,7,8,9), (10,11,12,13), (14,15,16,17), (18,19,20,21), (2,6,10,14,18), (3,7,11,15,19), (4,8,12,16,20), (5,9,13,17,21), (1,2,3,4,5,6,7,8,9,10,11,12,13,14,15,16,17,18,19,20,21)	& All joints\\
\noalign{\smallskip}\hline
\end{tabular}
 }
\end{center}
\end{table*}

\subsection{Datasets}
\begin{figure}[htbp]
\centerline{\includegraphics[width=6.7cm, height=9cm]{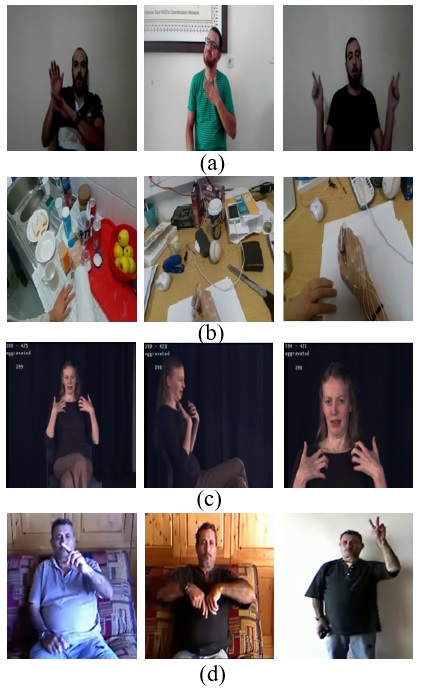}}
\caption{Dataset samples: (a) RKS-PERSIANSIGN, (b) First-Person, (c) ASLVID, (d) isoGD.}
\label{fig 3}
\end{figure}

Four datasets, \cite{b5,First-Person,ASLVID,isoGD}, are used in our evaluations. Details can be found in Table \ref{Table 3}. Only the RGB samples of the First-Person and isoGD datasets are used in our evaluations. Furthermore, some samples of these datasets are shown in Figure \ref{fig 3}. \\

\begin{table*}[h!]
\thispagestyle{empty}
\caption{\label{Table 3} Details of the dataset samples used in the train and test. In this table, CN is used for Class Number.}
\begin{center}
{\small
 \noindent\begin{tabular}{p{3.3cm}p{2cm}p{2cm}p{1.5cm}p{1.5cm}p{1.5cm}p{1.5cm}}
 \hline
\textbf{Dataset} & \textbf{Total samples} & \textbf{Total CN} & \multicolumn{2}{c}{\textbf{First protocol}} & \multicolumn{2}{c}{\textbf{Second Protocol}}\\
\hline\hline
&&& Train CN & Test CN & Train CN & Test CN\\
\hline\hline
RKS-PERSIANSIGN \cite{b5} & 10000 & 100 & 80 & 20 & 50 & 50\\
First-Person \cite{First-Person} & 10000 & 45 & 36 & 9 & 22 & 22\\
ASLVID \cite{ASLVID} & 9,800 & 250 & 200 & 50 & 125 & 125\\
isoGD \cite{isoGD} & 47933 & 249 & 199 & 50 & 124 & 124\\
\noalign{\smallskip}\hline
\end{tabular}
 }
\end{center}
\end{table*}

\subsection{Experimental results}
Here, we report the results of the proposed model using the two protocols mentioned in the previous subsection.\\

\subsubsection{Ablation analysis}	
We analyze the impact of different configurations of the proposed model.\\ 
\textbf{Different visual features:} We use both of the skeleton-based and deep features in the proposed model. To analyze the impact of each feature, we performed a detailed analysis on each feature, shown in Table \ref{Table 4}. As this table shows, the proposed model has the highest performance using the fused features.

\begin{table*}[h!]
\thispagestyle{empty}
\caption{\label{Table 4} Recognition accuracy of the proposed model using different features combination.}
\begin{center}
{\small
 \noindent\begin{tabular}{p{2.3cm}p{5cm}p{0.8cm}p{0.8cm}p{0.8cm}p{0.8cm}p{0.8cm}p{0.8cm}p{0.8cm}p{0.8cm}}
 \hline
\textbf{Visual Modality} & \textbf{Method} & \multicolumn{2}{c}{\textbf{RKS-PERSIANSIGN}} & \multicolumn{2}{c}{\textbf{First-Person}} & \multicolumn{2}{c}{\textbf{ASLVID}} & \multicolumn{2}{c}{\textbf{isoGD}}\\
\hline\hline
& & P1 & P2 & P1 & P2 & P1 & P2 & P1 & P2 \\
\hline\hline
Skeleton & Distances + BERT & 52.4 & 50.8 & 50.2 & 48.6 & 54.0 & 46.0 & 46.5 & 40.0	\\
& Angles + BERT	& 52.8 & 51.2 & 51.6 & 49.2 & 54.2 & 46.1 & 46.8 & 40.2\\
& SVD + BERT & 53.4 & 52.0 & 52.5 & 50.1 & 54.8 & 46.3 & 47.2 & 41.0\\
& Distances + Angles + BERT & 53.1 & 51.8 & 52.1 & 49.8 & 55.1 & 47.8 & 47.0 & 40.8\\
& SVD + Angles + BERT & 54.0 & 52.3 & 53.7 & 50.7 & 57.0 & 49.4 & 48.1 & 41.6\\
& SVD + Distances + BERT	& 54.2 & 52.6 & 54.1 & 51.1 & 58.2 & 50.8 & 48.8 & 42.5\\ 
& Distances + Angles + SVD + BERT & 55.4 & 53.1 & 54.9 & 51.9 & 58.8 & 51.4 & 49.4 & 43.6\\
\hline
RGB & C3D + LSTM + BERT	& 59.7 & 55.2 & 55.5 & 52.5 & 60.1 & 52.6 & 50.2 & 44.2\\
\hline
Skeleton+RGB & Distances + C3D + LSTM + BERT & 64.2 & 57.8	& 57.6 & 53.2 & 61.5 & 53.6 & 50.9 & 44.9\\
& Angles + C3D + LSTM + BERT & 65.4 & 59.3 & 58.8 & 53.9 & 62.2 & 54.8	& 51.8 & 45.8\\								
& SVD + C3D + LSTM + BERT & 68.1 & 62.1 & 61.4 & 54.7 & 63.1 & 55.7 & 52.5 & 46.6\\	
& Angles + SVD + C3D + LSTM + BERT & 68.8 & 62.8 & 61.8 & 55.0 & 63.6 & 55.8 & 53.8 & 46.8\\								
& Distances + SVD + C3D + LSTM + BERT & 69.4 & 63.4 & 62.2 & 55.2 & 64.0 & 56.1 & 54.1 & 47.2\\								
& Distances + Angles + C3D + LSTM + BERT	& 66.2 & 61.1 & 60.8 & 54.8 & 61.6 & 55.6 & 53.6 & 46.6\\ 							
& \textbf{Distances + Angles + SVD + C3D + LSTM + BERT} & \textbf{74.6} & \textbf{69.2} & \textbf{67.2} & \textbf{61.7} & \textbf{68.8} & \textbf{60.5} & \textbf{60.2} & \textbf{52.1}\\
\noalign{\smallskip}\hline
\end{tabular}
 }
\end{center}
\end{table*}

\textbf{Different input modalities:} We initially included only the skeleton modality in the model. Skeleton representation can be beneficial since it is compact, low-complex, and robustly separate the subject (human) from the background. However, obtaining higher performance using only one modality is hard. In this way, we included the pixel information corresponding to the RGB video modality. The deep visual features are obtained using the C3D model belonging to the LSTM network. Using two modalities in the visual embedding can provide complementary capabilities to the model. Results of the model using these modalities are shown the Table \ref{Table 4}. As this table shows, the proposed model obtains the highest performance using the fused features of the skeleton and video modalities.

\textbf{LSTM network:} Using different hidden neurons in the LSTM network can change the model performance. We used different hidden numbers in the LSTM network and reported the results in Table \ref{Table 5}. As this table shows, the proposed model obtains a higher performance using 1024 hidden neurons in the LSTM network.

\begin{table*}[h!]
\thispagestyle{empty}
\caption{\label{Table 5} Recognition accuracy of the proposed model using different hidden layers for LSTM network. In this table, LSTM-N indicates an LSTM network with N hidden neurons.}
\begin{center}
{\small
 \noindent\begin{tabular}{p{5cm}p{0.8cm}p{0.8cm}p{0.8cm}p{0.8cm}p{0.8cm}p{0.8cm}p{0.8cm}p{0.8cm}}
 \hline
\textbf{Model} & \multicolumn{2}{c}{\textbf{RKS-PERSIANSIGN}} & \multicolumn{2}{c}{\textbf{First-Person}} & \multicolumn{2}{c}{\textbf{ASLVID}} & \multicolumn{2}{c}{\textbf{isoGD}}\\
\hline\hline
& P1 & P2 & P1 & P2 & P1 & P2 & P1 & P2 \\
\hline\hline

LSTM-256 & 72.7 & 67.1	& 66.1 & 59.8 & 67.1 & 59.1 & 58.3 & 50.6\\
LSTM-512 & 73.4 & 68.8 & 66.3 &	60.2 & 67.9 & 59.9 & 59.4 & 51.4\\
LSTM-1024 & \textbf{74.6} & \textbf{69.2} & \textbf{67.2} & \textbf{61.7} & \textbf{68.8} & \textbf{60.5} & \textbf{60.2} & \textbf{52.1}\\
\noalign{\smallskip}\hline
\end{tabular}
 }
\end{center}
\end{table*}

\textbf{Deep network in the semantic space:} To project the visual features into the lingual embedding, Deep Neural Networks (DNNs) can be used due to their capabilities in the function approximation. Relying on the learning process, DNNs can learn the mapping patterns from the visual features into the lingual embedding. After the training of DNN, this model can be used to predict the lingual embedding for the unseen visual features. We used different deep networks with different settings. Finally, we selected a deep model including two dense layers (see Table \ref{Table 6}).

\begin{table*}[h!]
\thispagestyle{empty}
\caption{\label{Table 6} Recognition accuracy of the proposed model using different configurations of the semantic space. In this table, LSTM-N-M indicates an LSTM network with N hidden neurons connected to a DNN  with  M FC layers.}
\begin{center}
{\small
 \noindent\begin{tabular}{p{6cm}p{0.8cm}p{0.8cm}p{0.8cm}p{0.8cm}p{0.8cm}p{0.8cm}p{0.8cm}p{0.8cm}}
 \hline
\textbf{Model} & \multicolumn{2}{c}{\textbf{RKS-PERSIANSIGN}} & \multicolumn{2}{c}{\textbf{First-Person}} & \multicolumn{2}{c}{\textbf{ASLVID}} & \multicolumn{2}{c}{\textbf{isoGD}}\\
\hline\hline
& P1 & P2 & P1 & P2 & P1 & P2 & P1 & P2 \\
\hline\hline
LSTM-256-1 & 72.3 & 66.8 & 65.6 & 59.2 & 66.4 & 58.8	& 57.9 & 50.1\\
LSTM-256-2 & 72.7 & 67.1 & 66.1 & 59.8 & 67.1 & 59.1 & 58.3 & 50.6\\
Ours (LSTM-512-1) &73.1 & 68.4 & 66.0 & 60.0 & 67.4 & 59.5 & 59.1 & 51.0 \\
Ours (LSTM-512-2) &73.4 & 68.8 & 66.3 & 60.2 & 67.9 & 59.9 & 59.4 & 51.4 \\
Ours (LSTM-1024-1) &73.8 & 69.0 & 66.8 & 60.9 & 68.1 & 60.1 & 59.8 & 51.9 \\
Ours (LSTM-1024-2) & \textbf{74.6} & \textbf{69.2} & \textbf{67.2} & \textbf{61.7} & \textbf{68.8} & \textbf{60.5} & \textbf{60.2} & \textbf{52.1}\\
\noalign{\smallskip}\hline
\end{tabular}
 }
\end{center}
\end{table*}

\textbf{Different hand detection models:} 
Due to the high importance of the hand detection module in the SLR, we perform an analysis on the most used models for hand detection, Faster-RCNN, SSD, YOLO, and Transformer, in the proposed model. As Table \ref{Table 7} shows, our model achieves a higher accuracy using the Transformer model for hand detection.

\begin{table*}[h!]
\thispagestyle{empty}
\caption{\label{Table 7} Comparison of different hand detection models used in the proposed model.}
\begin{center}
{\small
 \noindent\begin{tabular}{p{4cm}p{0.8cm}p{0.8cm}p{0.8cm}p{0.8cm}p{0.8cm}p{0.8cm}p{0.8cm}p{0.8cm}}
 \hline
\textbf{Hand detection model} & \multicolumn{2}{c}{\textbf{RKS-PERSIANSIGN}} & \multicolumn{2}{c}{\textbf{First-Person}} & \multicolumn{2}{c}{\textbf{ASLVID}} & \multicolumn{2}{c}{\textbf{isoGD}}\\
\hline\hline
& P1 & P2 & P1 & P2 & P1 & P2 & P1 & P2 \\
\hline\hline
Faster-RCNN & 70.2 & 67.2 & 65.1 & 58.8 & 66.1 & 58.3 & 58.2 & 49.2 \\
SSD & 71.9 & 68.6 & 66.2 & 60.1 & 67.2 & 59.3 & 59.1 & 50.9 \\
YOLO & 72.1 & 68.9 & 66.9 & 60.8 & 67.9 & 59.9 & 59.6 & 51.3 \\
Transformer & \textbf{74.6} & \textbf{69.2} & \textbf{67.2} & \textbf{61.7} & \textbf{68.8} & \textbf{60.5} & \textbf{60.2} & \textbf{52.1}\\
\noalign{\smallskip}\hline
\end{tabular}
 }
\end{center}
\end{table*}

\subsubsection{Comparison with state-of-the-art models}
We compare our results with the state-of-the-art alternatives in the ZS-SLR (See Table \ref{Table 7}). We followed our evaluation protocols described in the previous sub-sections. Furthermore, we report the results of the proposed model after averaging on ten runs. In each run, we randomly select the training and testing classes. Since there is only one work in the ZS-SLR, we only compare the proposed model with Bigle et al. \cite{Bilge}. As Table \ref{Table 8} shows, the proposed model outperforms the state-of-the-art model in ZS-SLR.\\

\begin{table*}[h!]
\thispagestyle{empty}
\caption{\label{Table 8} Comparison with state-of-the-art models}
\begin{center}
{\small
 \noindent\begin{tabular}{p{4cm}p{0.8cm}p{0.8cm}p{0.8cm}p{0.8cm}p{0.8cm}p{0.8cm}p{0.8cm}p{0.8cm}}
 \hline
\textbf{Model} & \multicolumn{2}{c}{\textbf{RKS-PERSIANSIGN}} & \multicolumn{2}{c}{\textbf{First-Person}} & \multicolumn{2}{c}{\textbf{ASLVID}} & \multicolumn{2}{c}{\textbf{isoGD}}\\
\hline\hline
& P1 & P2 & P1 & P2 & P1 & P2 & P1 & P2 \\
\hline\hline
\cite{Bilge} & - & - & - & - & 51.4 & - & - & -\\			
Ours & \textbf{74.6} & \textbf{69.2} & \textbf{67.2} & \textbf{61.7} & \textbf{68.8} & \textbf{60.5} & \textbf{60.2} & \textbf{52.1}\\				
\noalign{\smallskip}\hline
\end{tabular}
 }
\end{center}
\end{table*}

\section{Discussion}
We analyze the proposed model as follows:\\
\textbf{ZS-SLR:} Although many models have been proposed and obtained the state-of-the-art performance in SLR \cite{b1,b2,b3,b4,b5,s1,s2}, they suffer from the annotation bottleneck and do not work efficiently for unseen classes. To overcome the mentioned weakness, we formulated ZS-SLR with no annotated visual examples. We performed different analysis of the proposed model to provide a basis for further exploration of ZS-SLR problem.

\textbf{Hand detection:} Since hand detection is an important step in SLR, we analyzed some of the most-used models, such as Faster-RCNN, SSD, and YOLO, for hand detection. The performance of these models is highly dependent on the hand-based components, such as a NMS method or anchor production. These components need to explicitly encode the prior knowledge about the task. To tackle these challenges, we configured the DEtection TRansformer (DETE) model, as a Transformer-based model for object detection, for hand detection in the our model and obtained a higher performance compared to other hand detection models.

\textbf{Hybrid features representation:} We fused some skeleton-based features, including the distances, angles, and singular values from SVD, with deep features to obtain more discriminative visual features. The step-by-step analysis of different features used in the model showed that the model obtained the higher performance using the fused features (See Table \ref{Table 4}). In the first analysis, only one type of skeleton-based feature was used in the model. As the first three rows of Table \ref{Table 4} show, the model performance is higher using the SVD features. This leads to assign a higher weight to the SVD features compared to the other skeleton-based features. In another analysis, the pixel-level features were included. We firstly analyzed the effect of the pixel-level, after that, one or more skeleton-based features were fused with the pixel-level features. As the last seventh rows of Table \ref{Table 4} show, the proposed model is more accurate relying on the complementary capabilities of the fused features.

\textbf{Multi-modal inputs:} Two visual modalities, the pixel-level information in the RGB video input and the compact representation of the hand skeleton, were fused to make a more powerful projection from the visual features into the lingual embedding obtained from the third modality. Relying on the results, our model successfully benefited from the complementary capabilities of the fused features from multi-modal inputs and improved the state-of-the-art results in ZS-SLR.

\begin{figure*}[htp]
\centerline{\includegraphics[width=0.65\linewidth]{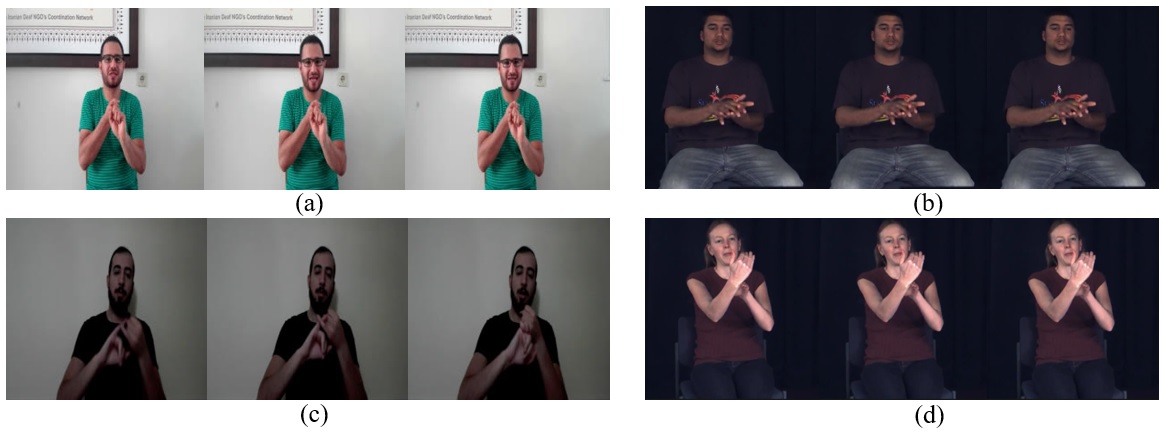}}
\caption{False recognition samples from the RKS-PERSIANSIGN (a, c) and ASLVID datasets (b, d): (a) Predicted: Sister, True label: Brother, (b) Predicted: Bathing suit, True label: Alligator, (c) Predicted: Brother, True label: Sister, (d) Predicted: Algebra, True label: Sailboat.}
\label{fig 4}
\end{figure*}

\begin{figure*}[htp]
\centerline{\includegraphics[width=0.65\linewidth, height=5.5cm]{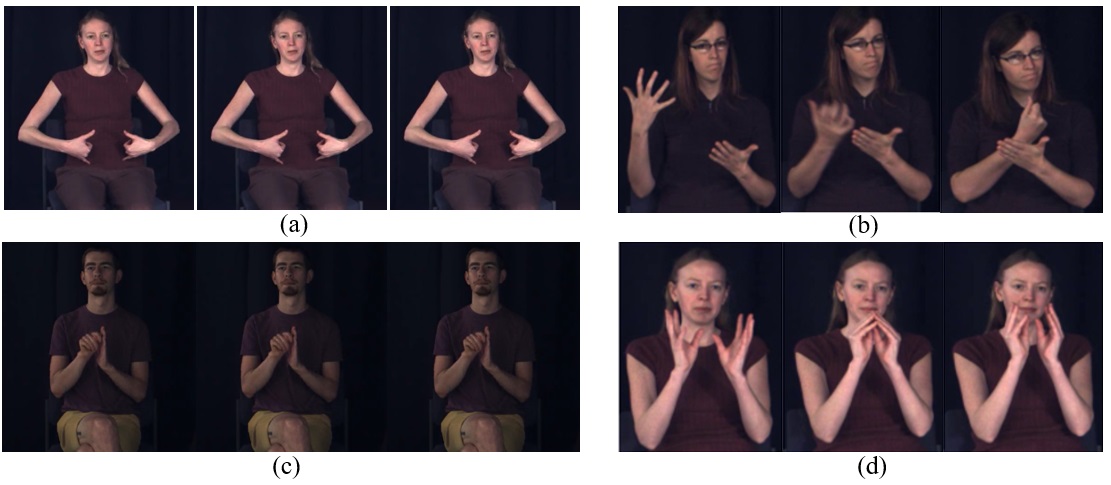}}
\caption{True recognition samples from the ASLVID dataset: (a) Bathing suit, (b) Alligator, (c) Old testament, (d) Sailboat.}
\label{fig 5}
\end{figure*}

\begin{figure*}[htp]
\centerline{\includegraphics[width=0.65\linewidth]{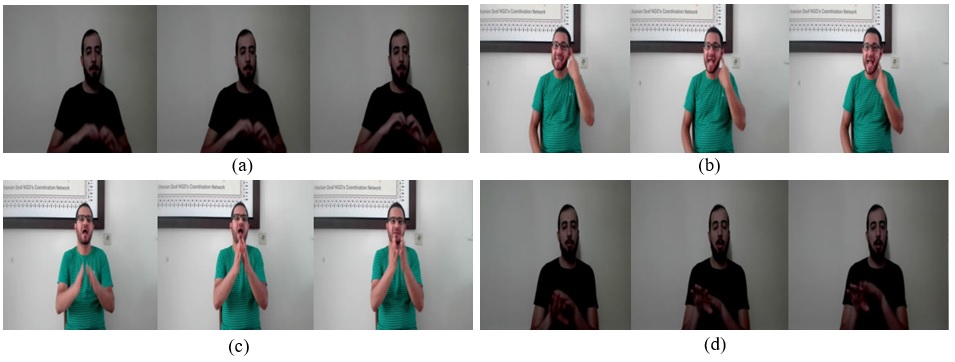}}
\caption{True recognition samples from the RKS-PERSIANSIGN dataset: (a) Love, (b) Woman, (c) Book, (d) Kind.}
\label{fig 6}
\end{figure*}

\textbf{Performance:} The step-by-step analysis of the proposed model shows the outperforming of the model performance compared to state-of-the-art alternatives in the ZS-SLR. The proposed model decreased the false recognition using the complementary capabilities of the hybrid features and multi-modalities. This means that if the model cannot have a powerful capability in each of the features or modalities, it has this chance to boost the performance by relying on the other features or modalities. This makes the model more robust and useful because the model is not biased to a special feature or modality. However, there is much room to improve the recognition accuracy in the ZS-SLR. Since our main focus in this work was on the sign language, we provided more analysis on the false recognition samples in SLR datasets, RKS-PERSIANSIGN and ASLVID. Analyzing the results on these datasets showed that the proposed model obtained a higher performance on the RKS-PERSIANSIGN. Results showed that the recognition accuracy of each sign in the RKS-PERSIANSIGN dataset is more than 0.7. Fig. \ref{fig 4}, \ref{fig 5}, and \ref{fig 6} show some samples of false and true recognition in the sign datasets. As one can see in Fig. \ref{fig 4}, some signs, such as 'Sister', 'Brother', 'Alligator', and 'Sailboat' are challenging and difficult to discriminate from the other signs. Analyzing the false recognition showed that there are some similarities between these signs and some signs in these datasets. Therefore, increasing the samples of these signs can reduce miss-classifications and prepare the model to learn more powerful discriminative patterns in these categories. Complex patterns in different signs in the real environment using unseen data during the training is hard and challenging. Compared to the related areas, such as action recognition, in SLR, even a soft variation in motion and/or handshape can variate the whole meaning. Thus, dedicated models are indispensable for ZS-SLR.\\ 

%

\section{Conclusion and future work}
In this work, we proposed a multi-modal deep learning-based model for ZS-SLR. In this model, a Transformer-based model was configured for hand detection. The proposed model successfully benefited from the complementary capabilities of deep features fused with skeleton-based features. Three skeleton-based features, containing the distances, angles, and singular values obtained from the SVD method, were fused. The pixel-level features obtained from the C3D model were employed in the deep features to input to the LSTM network. Furthermore, an AE model was fell on the LSTM output to make a dimension reduction on the deep features and balance the dimensionality of the skeleton-based and deep features. The fused features were used in the semantic space to map the visual features into the lingual embedding obtained from the BERT model. We performed a detailed analysis of the model and reported the results. While the proposed model achieved state-of-the-art results in ZS-SLR on four large-scale datasets, RKS-PERSIANSIGN, First-Person, ASLVID, and isoGD, some challenges of ZS learning still need to be addressed. A powerful discriminative model for semantic space is a key part of the ZS model. We would like to check the effectiveness of substituting the CNN model with the Transformer model. This refers to the limitations of the CNN model in terms of the small receptive field that it can capture.\\


\vspace{12pt}

\end{document}